\pdfoutput=1

\documentclass[11pt]{article}

\usepackage[preprint]{coling}

\usepackage{times}
\usepackage{latexsym}

\usepackage[T1]{fontenc}

\usepackage[utf8]{inputenc}

\usepackage{microtype}
\usepackage{url}
\usepackage{inconsolata}

\usepackage{graphicx}

\usepackage{booktabs}
\usepackage{multirow} 
\title{Refining Translations with LLMs: A Constraint-Aware Iterative Prompting Approach}

\author{
 \textbf{Shangfeng Chen\textsuperscript{1}},
 \textbf{Xiayang Shi\textsuperscript{1*}},
 \textbf{Pu Li\textsuperscript{1}},
 \textbf{Yinlin Li\textsuperscript{2}},
 \textbf{Jingjing Liu\textsuperscript{1}}
\\
 \textsuperscript{1}Zhengzhou University of Light Industry
 \\
 \textsuperscript{2}Institute of Automation, Chinese Academy of Sciences
\\
 \small{
   \textbf{Correspondence:} \href{mailto:aryang123@163.com}{aryang123@163.com}
 }
}

\begin{document}
\maketitle
\begin{abstract}

Large language models (LLMs) have demonstrated remarkable proficiency in machine translation (MT), even without specific training on the languages in question. 
However, translating rare words in low-resource or domain-specific contexts remains challenging for LLMs. 
To address this issue, we propose a multi-step prompt chain that enhances translation faithfulness by prioritizing key terms crucial for semantic accuracy. 
Our method first identifies these keywords and retrieves their translations from a bilingual dictionary, integrating them into the LLM's context using Retrieval-Augmented Generation (RAG). 
We further mitigate potential output hallucinations caused by long prompts through an iterative self-checking mechanism, where the LLM refines its translations based on lexical and semantic constraints. 
Experiments using Llama and Qwen as base models on the FLORES-200 and WMT datasets demonstrate significant improvements over baselines, highlighting the effectiveness of our approach in enhancing translation faithfulness and robustness, particularly in low-resource scenarios.

\end{abstract}

\section{Introduction}
\noindent


The rapid advancement of large language models (LLMs) has profoundly impacted the field of machine translation (MT).  
Recent studies have demonstrated the remarkable capacity of LLMs to perform zero-shot and few-shot translation tasks, achieving impressive results even without explicit training on parallel corpora \cite{zhu2024towards,zhu2024landermt}. 
This inherent multilingual capability, arising from vast pre-training on diverse textual data, presents new opportunities for low-resource MT, a domain traditionally hindered by data scarcity \cite{robinson-etal-2023-chatgpt}.

However, despite their promising potential, LLMs still struggle to consistently achieve the faithfulness and precision required for high-quality translation, especially when dealing with rare or specialized terminology \cite{cui2024efficiently}.
LLMs are typically pre-trained on data that exhibits a significant skew towards high-resource languages, leading to an under-representation of linguistic nuances and vocabulary associated with low-resource languages. 
This bias results in a performance gap when translating between or into low-resource languages \cite{2023arXivDictionary-base}.
Rare words often exhibit a higher degree of polysemy, making it challenging for LLMs to accurately discern the correct translation based on context.
This ambiguity can lead to semantically inaccurate or misleading translations \cite{robinson-etal-2023-chatgpt}.

Several studies have explored the integration of external knowledge into the LLM-based MT pipeline. 
One prominent approach involves incorporating bilingual dictionaries or lexical constraints into the prompting process \cite{2023arXivDictionary-base,lu2023chain-chain-of-dictionary}. 
While these efforts have shown promise in improving translation accuracy, they often rely on simply appending pre-translated terms to the prompt, neglecting a more nuanced analysis of the source sentence. 
Furthermore, existing methods often overlook the potential for leveraging LLMs themselves as agents for both translation and refinement. 

To address this issue, we propose a multi-step prompt chain that aims to enhance the faithfulness and robustness of LLM-based MT.
In details, we first identifies and extracts keywords in the source sentence critical to translation quality. 
Then, we apply retrieval-augmented generation (RAG) \cite{RAG} to embed translations of these keywords, retrieved from a bilingual dictionary, into the LLM’s context. 
Finally, we refine translations through iterative revisions by employing LLMs as constraint-aware translators and revisers.
Through comprehensive experiments on the FLORES-200 benchmark dataset for low-resource languages \cite{nllb2022} and contamination-free WMT datasets \cite{arthur-etal-2016-incorporating,duan2020bilingual,zhong2020retracted}, we demonstrate that our method significantly outperforms existing approaches, achieving state-of-the-art results on multiple language pairs.



Our main contributions are three-fold:

\begin{itemize}
    \item The proposed method allows for the effective identification of keywords that are crucial to the quality of translation, while simultaneously filtering out less relevant words that do not contribute to accuracy. Furthermore, the method ensures adherence to the constraints that are typically associated with translation.
    
    \item The proposed approach employs Retrieval-Augmented Generation (RAG) to improve translation quality. By incorporating keywords retrieved from bilingual dictionaries into the LLM's context window and applying a post-translation self-checking mechanism, the method minimizes the potential for misunderstandings and optimizes.
    
    \item We further advanced machine translation tasks employing LLMs with 7B parameters by leveraging prompt-based techniques, obviating the necessity for fine-tuning. This approach yielded notable improvements in translation performance.
    
\end{itemize}

\section{Related Work}
\subsection{Prompt of LLM for MT} 
\noindent

Prompting techniques has grown in popularity as a means of harnessing the capabilities of large language models in the field of machine translation. 
Prior research \cite{10.5555/3495724.3495883} has demonstrated that LLMs can perform zero-shot and few-shot translation tasks effectively by using simple prompts, even in the absence of parallel corpora. 
These initial findings established a foundation for further investigation into the potential of prompt-based approaches in machine translation \cite{zhu2024feds}. 
\citet{vilar-etal-2023-prompting} expanded on this by using prompts to control various aspects of translation outputs, such as formality and dialect. 
Their work underscored the potential of prompting to customize translations in accordance with particular requirements. 
\citet{Garcia-unreasonable-effectiveness} examined the unreasonable effectiveness of few-shot learning in MT, emphasizing the role of prompt engineering in achieving high-quality translations even in low-resource scenarios. 
Moreover, \citet{lin-etal-2022-few-shot} and \citet{agrawal-etal-2023-context} have focused on selecting effective in-context examples to enhance translation performance. \citet{2023arXivDictionary-base} and \citet{lu2023chain-chain-of-dictionary} introduced dictionary-based phrase-level prompting, showcasing how bilingual dictionaries can be leveraged within prompts to guide LLMs in translating rare words and phrases more accurately. 
\citet{sarti-etal-2023-ramp} proposed RAMP (Retrieval and Attribute-Marking Enhanced Prompting), which improves attribute-controlled translation by integrating retrieval-augmented prompts with LLMs, thereby leading to more contextually accurate translations.

\subsection{Lexical-based in MT}
\noindent

Lexical-based approaches in MT have received  considerable attention, particularly with regard to the enhancement of translation quality for rare or domain-specific terms. \citet{zhang2016bridging} presents a technique that enhances NMT by incorporating a bilingual dictionary, specifically targeting less common or previously unseen words found in the bilingual training data.
\citet{lample-etal-2018-phrase} employed bilingual dictionaries, in unsupervised MT, has been explored. 
Their findings indicate that the incorporation of dictionary entries into the MT pipeline can significantly improve translation quality, particularly for low-resource languages \cite{mi2021improving}. 
\citet{nmt-dictionary2020} utilize a dictionary to create synthetic parallel data, which is then used to train a more resilient NMT model. In a related study, \citet{lin2021bilingual} incorporated the translation information from dictionaries into the pre-training process and propose a novel Bilingual Dictionary-based Language Model (BDLM). This approach has been demonstrated to be effective in the context of low-resource language translation.

\begin{figure*}[ht]
  \centering
  \includegraphics[width=1.0\textwidth]{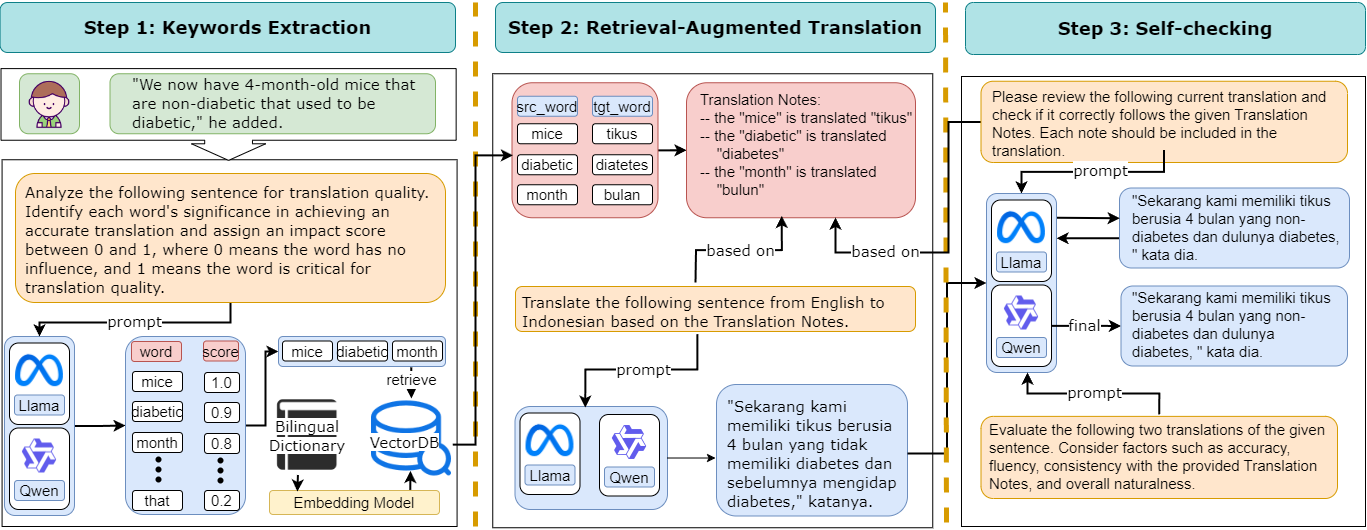}
  \caption{The proposed method of translation process.}
  \label{fig1}
\end{figure*}

\subsection{Reflection on LLM}
\noindent

Reflection mechanisms in LLMs has emerged as a prominent area of investigation, with the objective of enhancing the dependability and precision of the generated outputs. \citet{wang2022self} introduced a self-consistency technique in which models generate multiple solutions and select the most consistent one, demonstrating significant improvements in reasoning tasks.
\citet{RAG_self} proposed a retrieval-augmented generation (RAG) framework that allows LLMs to verify their generated content by retrieving external knowledge, which is particularly beneficial for knowledge-intensive tasks. Similarly, \citet{asai2023self} proposed Self-RAG which LLMs generates an initial output and then retrieves relevant information to verify and refine their own response, thereby improving accuracy and consistency in knowledge-intensive tasks. 
In the context of fact-checking, \citet{lee-etal-2020-language} integrated self-verification capabilities into LLMs, enabling them to cross-check their factual outputs. \citet{reflexion} explored a framework that improves language agents' decision-making by using linguistic feedback rather than model updates, enabling significant performance gains across diverse tasks without extensive fine-tuning.

\section{Methodology}
\noindent

This section outlines the core principles of our proposed methodology and its implementation details, offering a clear overview of the technique and the rationale behind its design.
As illustrated in Figure \ref{fig1}, our method differs by focusing on identifying and extracting keywords from the source text and embedding them into the translation process via structured prompts.

\subsection{Keywords Extraction}
\noindent

In MT, accurately translating keywords such as domain-specific or low-frequency terms, is  crucial for preserving semantic integrity. 
These words have a significant impact on translation quality, and errors can lead to ambiguity or loss of meaning. 
However, LLMs often focus on sentence sequences, neglecting  the importance of these key terms. Our approach emphasizes the identification and extraction of keywords to enhance translation accuracy.

\begin{figure}[h]
    \centering
    \includegraphics[width=1.0\linewidth]{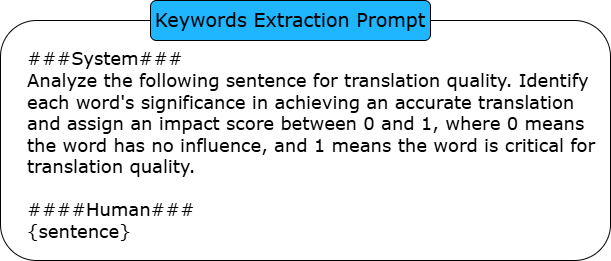}
    \caption{The prompt template for keywords extraction.}
    \label{keywords extraction}
\end{figure}

The first step in our approach involves analyzing the source sentence to identify the words that are most critical for accurate translation. The keywords extraction prompt template is shown in Figure \ref{keywords extraction}. 

We hypothesize that certain words, due to their semantic weight or specificity, play a more crucial role in preserving the intended meaning of the sentence. Given a source sentence $X=\{x_1,x_2,\dots,x_n\}$, The LLM generates a priority score $P=\{p_1,p_2,\dots,p_n\}$, as:
\begin{equation}
    p_i=f_{LLM}(x_i \vert X)
\end{equation}
where $s_i$ represents individual words, our objective is to generate a priority score $p_i$ for each word $s_i$ in $S$. This score reflects the importance of the word in determining the quality of the translation. $f_{LLM}$ is the function implemented by the LLM, which takes the entire sentence $S$ and outputs the priority score $p_i$ for each word $s_i$. The LLM processes the sentence and outputs a sequence of priority scores $\{p_1,p_2,\dots,p_n\}$, where each $p_i \in [0,1]$ indicates the importance of word $s_i$. The higher the score, the more critical the word is for maintaining translation fidelity.

Ultimately, based on the above sentence $X=\{x_1,x_2,\dots,x_n \}$ and priority score $P=\{p_1,p_2,\dots,p_n\}$, using $k$ as the number of top-scoring words to be selected. The goal is to select the subset $W$ of $k$ words from $S$ that have the highest scores in $P$. This can be mathematically described as:
\begin{equation}
    \mathcal{W}=\{w_i \vert w_i \in X, p_i \in Top_k(P) \}
\end{equation}
Here, $Top_k(P)$ denotes the set of the top $k$ scores from the score set $P$. The selected words $W$ are those whose scores belong to the top $k$ scores in the list. The value of $k$ is an adaptive threshold based on the length of the sentence.

\subsection{Retrieval-Augmented Translation}
\noindent

After extracting keywords from the sentence, the next step involves retrieving their corresponding translations from a bilingual dictionary.  Each identified keyword $w_i$ is represented as a vector $\mathbf{v}_{w_i}$ in the bilingual dictionary's vector space. 

The goal is to find the most accurate translation $t_i$ for each $w_i$ by searching the dictionary's vector database. This is achieved by embedding the word into a pre-trained vector space, such that:
\begin{equation}
    \mathbf{v}_{w_i}=\mathrm{Embeding}(w_i)
\end{equation}
where $\mathrm{Embeding}(\cdot)$ is the embedding function mapping the keyword $w_i$ to its vector representation.

The translation $t_i$ of the keyword $w_i$ is obtained by searching the nearest neighbor in the bilingual dictionary's vector space. This can be expressed as:
\begin{equation}
    t_i=\arg\min_{t \in \mathcal{T}} \mathrm{dist}(\mathbf{v}_{w_i},\mathbf{v}_t)
\end{equation}
where $\mathcal{T}$ represents the set of all translation candidates in the bilingual dictionary, and $\mathrm{dist}(\cdot,\cdot)$ is the distance metric used to measure the similarity between vectors, typically cosine similarity.

\begin{figure}[h]
    \centering
    \includegraphics[width=1.0\linewidth]{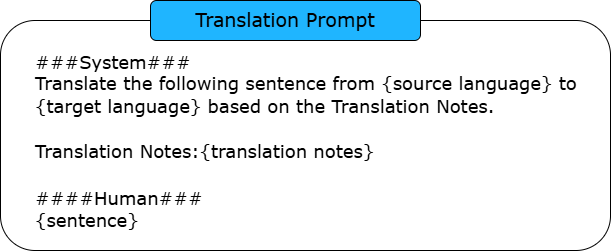}
    \caption{The prompt template for translation based on translation notes.}
    \label{fig2}
\end{figure}

Once the translation $t_i$ is retrieved, it will be fused with $w_i$ as a new constraint $\mathcal{C}=\{(w_1,t_1),(w_2, t_2),\dots, (w_k, t_k)\}$ into the context window of the prompt. Our prompt template is shown in Figure \ref{fig2}.

Let $Y^1$ denote the initial translation obtained by integrating the lexical constraints $\mathcal{C}$ into the LLM’s prompt during the first translation pass. 
This translation is produced by conditioning the LLM on both the source sentence $X$ and the aforementioned constraints $\mathcal{C}$:
\begin{equation}
    Y^1=f_{LLM}(X,\mathcal{C})
\end{equation}
where $Y^1=\{y_1^1, y_2^1, \dots, y_m^1\}$ represents the initial translation after incorporating the lexical constraints.

\subsection{Self-checking}
\noindent

\begin{figure}[h!]
    \centering
    \includegraphics[width=1.0\linewidth]{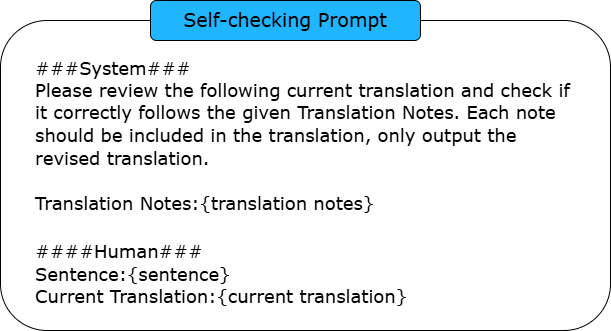}
    \caption{The prompt template for translation based on translation notes.}
    \label{self-checking}
\end{figure}

To ensure the effective utilization of all translation notes throughout the machine translation procedure, we implement an iterative prompting strategy. 
Although LLMs are capable of producing high-quality translations, they do not always incorporate all of the provided translation notes in a single iteration. 
Consequently, we have developed an iterative mechanism in which the translation process is repeated until all constraints are fully satisfied. The self-checking prompt template is shown in Figure \ref{self-checking}. 

After generating the initial translation $Y^1$ , the next step involves a self-checking mechanism that refines the translation. 
This self-checking step uses the current translation $Y^i=\{y_1^i,y_2^i,\dots, y_m^i\}$ and the lexical constraints $\mathcal{C}$ to improve the quality of the next translation $Y^{i+1}$. The self-checking process can be represented as:

\begin{equation}
    Y^{i+1}=f_{LLM}(X,\mathcal{C}, Y^i)
\end{equation}

The final translation $Y_{final}$ is generated after the self-checking and refinement steps. 
The LLM ensures that each keyword $w_i$ from the source sentence is translated into its corresponding target word $t_i$ from the constraints set $\mathcal{C}$, while maintaining the overall fluency and accuracy of the translation:

\begin{equation}
    Y_{final}=\{y_i \vert y_i=t_j, x_i=w_j, (w_j, t_j) \in \mathcal{C}\}
\end{equation}
Thus, the final output $Y_{final}$ is the refined translation, which incorporates both the initial lexical constraints and the improvements from the self-checking mechanism.

Finally, we designed a specific prompt to enable the model to autonomously evaluate which translation better satisfies the prescribed lexical constraints while preserving fluency and coherence.
The LLM evaluates both the initial translation $Y_1$ and the refined translation $Y_{final}$ then selects the one that offers superior quality based on these criteria. The evaluation process can be described mathematically as follows:
\begin{equation}
    Y_{best}=f_{LLM}(Y_1, Y_{final})
\end{equation}
This self-evaluation process allows the LLM to optimize translation output without human intervention. 

\section{Experimental Setup}
\noindent

\textbf{Models}
We employed a multi-step prompting approach for keywords extraction and translation, leveraging the strengths of different models. 
Specifically, we utilized the Mete-Llama-3.1-8B-Instruct model \cite{dubey2024llama} and the Qwen2-7B-Instruct \cite{qwen2} model for extracting keywords that have a significant impact on the quality of translation.
These models are chosen for their advanced natural language understanding capabilities, which enable them to accurately extract crucial words from the source sentences. 
For the embedding retrieval process, we used the bge-m3 \cite{bge-m3} model, which is specifically designed to handle large-scale embedding tasks. 
This model facilitates the mapping of identified keywords to their corresponding translations in a bilingual dictionary, thereby ensuring the maintenance of semantic integrity  throughout the translation process.

\textbf{Datasets}
While LLMs have shown exceptional performance in high-resource language tasks, we aim to assess their effectiveness in handling underrepresented languages, where data scarcity significantly challenges translation accuracy and quality. 
Our study focuses on evaluating the robustness and adaptability of LLMs in translating low-resource languages and those not encountered during training. 
We employed FLORES-200 devtest\cite{nllb2022}, which is a benchmark dataset for machine translation between English and low-resource languages. 
The creation of FLORES-200 doubles the existing language coverage of FLORES-101 \cite{flores101}. 
FLORES-200 consists of translations from 842 distinct web articles, totaling 3001 sentences. 
On average, sentences are approximately 21 words long.
As we aim to focus on low-resource languages, we selected 10 languages as Table \ref{tab1}.

\begin{table}[h!]
    \centering
    \resizebox{\linewidth}{!}{
    \begin{tabular}{ccc}
    \hline
    \textbf{ISO 639-1} & \textbf{Language} & \textbf{FLORES-200 code} \\
    \hline
    ca & \textbf{Catalan} & cat\_Latn \\
    hr & \textbf{Croatian} & hrv\_Latn \\
    da & \textbf{Danish} & dan\_Latn \\
    nl & \textbf{Dutch} & nld\_Latn \\
    tl & \textbf{Tagalog} & tgl\_Latn \\
    id & \textbf{Indonesian} & ind\_Latn \\
    it & \textbf{Italian} & ita\_Latn \\
    ms & \textbf{Malay} & zsm\_Latn \\
    nb & \textbf{Norwegian} & nob\_Latn \\
    sk & \textbf{Slovak} & slk\_Latn \\
    \hline
    \end{tabular}
}
\caption{The low-resource languages chosen from the FLORES-200 dataset, along with their ISO 639-1 code and FLORES-200 code.}
\label{tab1}
\end{table}

Meanwhile, \cite{zhou2023don} reported a issue that due to the fact that most of the corpus used for training large language models is publicly available, this indirectly leads to data pollution issues during the evaluation process of the model. 
To reduce this risk, we also evaluated the $EN \to DE$ language pairs from the WMT22, WMT23, and WMT24 test sets, which can effectively reduce the impact of data pollution.

\textbf{Evaluation Metrics}
For evaluation metrics, we reported the chrF++ \cite{popovic-2015-chrf} and the BLEU evaluations  provided by the sacreBLEU \footnote{\url{https://github.com/mjpost/sacrebleu}} \cite{sacreBLEU}. 
We employed MTME \footnote{\url{https://github.com/google-research/mt-metrics-eval}} which is a simple toolkit to evaluate the performance of Machine Translation metrics on WMT test sets.

\begin{table*}[h!]
    \centering
    \vspace{2mm}
    \begin{tabular}{c|cc|cc|cc|cc}
        \toprule
        \multirow{3}{*}{\textbf{Language}} & \multicolumn{4}{c|}{\textbf{Meta-Llama-3.1-8B-Instruct}} &  \multicolumn{4}{c}{\textbf{Qwen2-7B-Instruct} }\\
        \cline{2-9}
        & \multicolumn{2}{c|}{\textbf{BLEU}} & \multicolumn{2}{c|}{\textbf{chrf++}} & \multicolumn{2}{c|}{\textbf{BLEU}} & \multicolumn{2}{c}{\textbf{chrf++}} \\
        \cline{2-9}
        & \textbf{Baseline} & \textbf{Ours} & \textbf{Baseline} & \textbf{Ours} & \textbf{Baseline} & \textbf{Ours} & \textbf{Baseline} & \textbf{Ours} \\
        \hline
        ca-en & 41.48 & \textbf{41.95} & 66.09 & \textbf{66.63} & 39.18 & \textbf{42.12} & 64.14 & \textbf{65.38}  \\
        en-ca & 33.45 & \textbf{34.02} & 60.33 & \textbf{61.21} & 21.28 & \textbf{24.94} & 50.46  & \textbf{55.30}\\
        hr-en & \textbf{33.42} & 29.91 & \textbf{60.72} & 55.38 & 31.06 & \textbf{31.75} & 58.27 & \textbf{60.02} \\
        en-hr & 18.93 & \textbf{20.02} & 49.95 & \textbf{51.64} & 12.37 & \textbf{15.84} & 42.22 & \textbf{47.67} \\
        da-en & 43.74 & \textbf{44.24} & 67.70 & \textbf{67.91} & 41.99 & \textbf{42.38} & 66.04 & \textbf{66.94} \\
        en-da & 34.05 & \textbf{34.78} & 61.53 & \textbf{62.12} & 22.16 & \textbf{26.76} & 51.11 & \textbf{56.11} \\
        nl-en & 29.05 & \textbf{29.10} & 57.65 & \textbf{57.89} & \textbf{28.44} & 26.77 & \textbf{57.10} & 55.87 \\
        en-nl & 21.82 & \textbf{22.07} & 54.34 & 54.34 & 18.75 & \textbf{19.09} & 51.21 & \textbf{52.16} \\
        tl-en & 34.64 & \textbf{37.93} & 59.57 & \textbf{62.00} & 32.36 & \textbf{33.75} & 57.09 & \textbf{58.32} \\
        en-tl & 24.07 & \textbf{26.98} & 53.74 & \textbf{56.89} & 15.79 & \textbf{17.42}  & 44.43 & \textbf{47.89} \\
        id-en & 37.69 & \textbf{38.98} & 64.04 & \textbf{64.87} & 37.42 & \textbf{38.20} & 63.69 & \textbf{65.04} \\
        en-id & 36.54 & \textbf{36.95} & 63.37 & \textbf{64.00} & 32.89 & \textbf{33.55} & 60.82 & \textbf{62.14}\\
        it-en & 30.23 & \textbf{30.56} & 57.97 & \textbf{58.22} & 29.36 & \textbf{30.46}  & 57.38 & \textbf{59.46} \\
        en-it & 25.94 & \textbf{26.16} & 57.63 & \textbf{58.75} & 22.56 & \textbf{23.11} & 55.07 & \textbf{55.23}\\
        ms-en & \textbf{37.38} & 36.20 & \textbf{63.06} & 61.03 & 36.05 & \textbf{36.86} & 61.69 & \textbf{62.37}\\
        en-ms & 28.97 & \textbf{29.72} & 59.30 & \textbf{62.48} & 23.19 & \textbf{25.19} & 54.65 &  \textbf{57.20} \\
        no-en & \textbf{39.47} & 39.04 & \textbf{64.73} & 64.10 & 36.55 & \textbf{38.23} & 62.10 & \textbf{64.88} \\
        en-no & 25.32 & \textbf{25.43} & 55.60 & \textbf{56.11} & 17.22 & \textbf{21.07} & 47.33 & \textbf{52.81} \\
        sk-en & 33.04 & \textbf{33.80} & 60.14 & \textbf{60.79} & \textbf{30.99} & 29.14 & \textbf{58.47} & 56.40 \\
        en-sk & 17.40 & \textbf{20.53} & 46.91 & \textbf{49.18} & 11.16 & \textbf{14.52} & 39.73 & \textbf{45.35}\\
        \bottomrule
    \end{tabular}

    \caption{Comparing baseline and proposed method translation results for different low-resource to English and English to low-resource language pairs.}
    \label{tab2}
\end{table*}

\textbf{Dictionary}
We utilized the ground-truth bilingual dictionaries provided by \cite{dictionary}. These dictionaries command.\footnote{\url{https://github.com/facebookresearch/MUSE##ground-truth-bilingual-dictionaries}} were meticulously crafted using Meta’s internal translation tool, with specific attention to handling the polysemy of words, ensuring that multiple meanings of a word are accurately represented.

\textbf{Baseline}
For our baseline, we employ a straightforward prompting strategy where the LLMs are guided using the prompt \textit{``Translate the following sentence from \{source language\} to \{target language\}."} This prompt is tested in both zero-shot and few-shot settings. 
For few-shot evaluations, we randomly select examples from the development set to provide context. 
This approach allows us to establish a fundamental benchmark, enabling us to compare and assess the improvements introduced by our proposed method.

\section{Results and Analysis}
\noindent

In this section, we provide a detailed account of a series of experiments conducted for our proposed method, including Main Result, contamination-free evaluation and ablation studies aimed at comprehensively evaluating the effectiveness of our proposed approach.

\subsection{Main Results}
\noindent

Table \ref{tab2} presents a comparative analysis of the performance of our method relative to the baseline in the context of low-resource languages.
It is noteworthy that the Llama-3.1 model exhibited enhanced performance across the majority of language pairs, reflecting its augmented capacity to generate high-quality translations. 
However, a slight decline in performance was observed in a small subset of language pairs. 
The decline in performance for certain language pairs can be traced to the model's reliance on training data predominantly focused on Romance languages. 
In these cases, the model struggled with polysemy, leading to contextually incorrect translations. 
While our method alleviated some of these challenges, the limitations of the training data still exerted an influence, particularly when the model lacked exposure to diverse linguistic contexts. 
Despite this, our approach proved effective in enhancing the Meta-Llama-3.1-8B-Instruct model's performance across a wide range of languages.

For the Qwen2-7B-Instruct model, experimental results showed that our method yielded substantial improvements across nearly all language pairs compared to zero-shot translation. 
The primary issue arose from the hallucination of unintended Chinese content during zero-shot translation, even when the target language was not Chinese. 
This issue can be attributed to the substantial proportion of Chinese data included in the training process of the Qwen2 model.
The combination of lexical constraints with an iterative self-checking mechanism proved an effective approach to mitigating hallucinations, with a notable reduction in instances of erroneous content generation in non-target languages.

\begin{figure*}[htbp]
    \centering
    \includegraphics[width=0.9\linewidth]{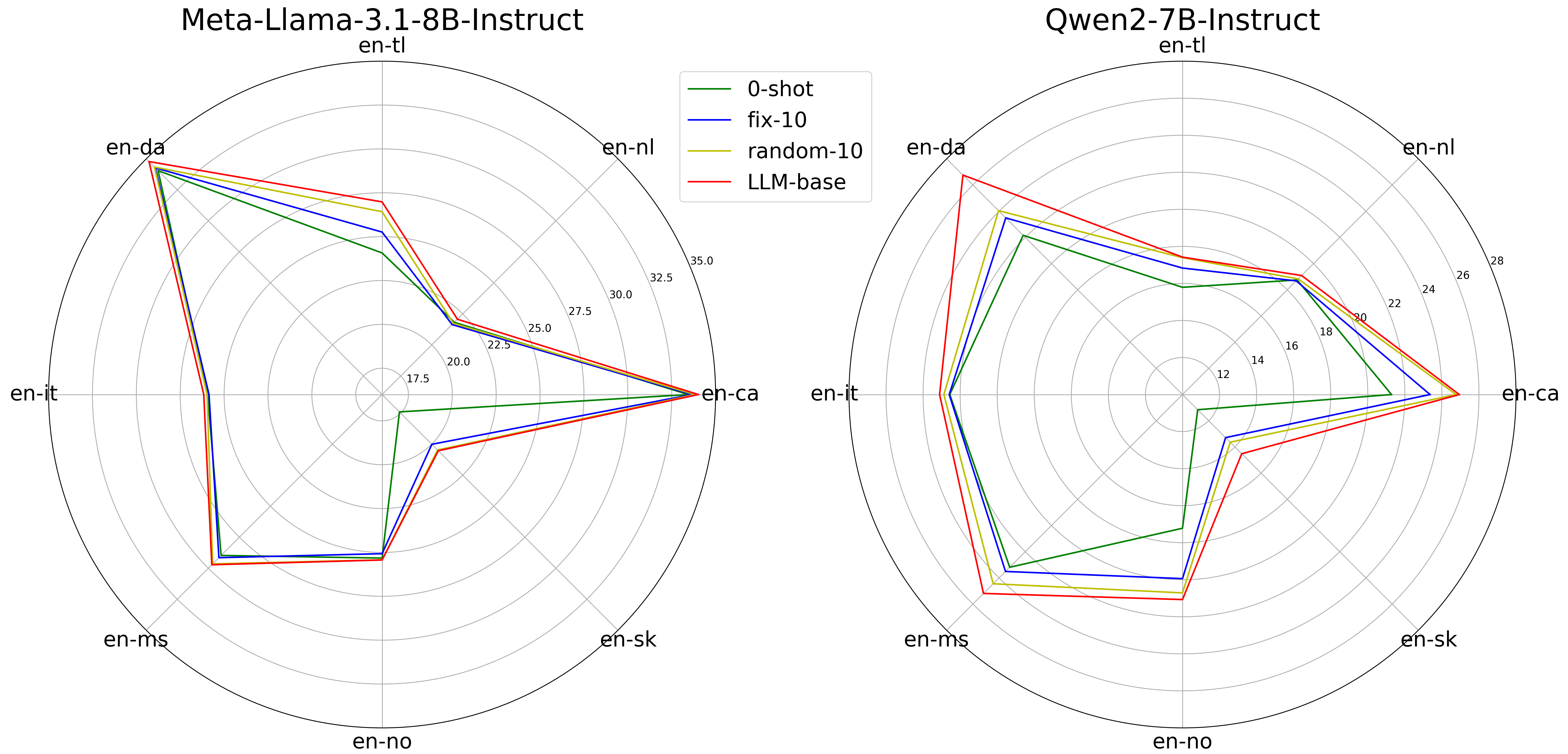}
    \caption{Comparison of BLEU scores across different word selection methods}
    \label{fig3}
\end{figure*}

The self-checking mechanism proved instrumental in mitigating translation hallucinations by virtue of its continuous refinement of the translation process and its assurance that the model remained faithful to the intended target language. 
This mechanism significantly reduced the occurrence of hallucinations, thereby improving overall translation accuracy. 
These findings highlight that the composition of the training data brings both advantages and challenges: while the inclusion of large-scale multilingual data enhanced overall performance, it also introduced hallucination issues in zero-shot scenarios. 
Our method effectively addressed these challenges, further demonstrating its potential for improving the translation quality of LLMs.

\subsection{Contamination-Free Evaluation}
\noindent

Moreover, experiments were conducted using the most recent machine translation evaluation datasets, specifically WMT22, WMT23, WMT24. 
These datasets provide a valuable benchmark for evaluating the performance of LLMs in translation tasks. 
One of the key advantages of utilizing these updated datasets is their ability to mitigate the issue of data contamination that can arise during the training process. 
Data contamination refers to the inadvertent encounter of portions of the test set by models during training, which can result in performance scores that are artificially inflated and do not accurately reflect the model's true capabilities in a real-world setting. 

\begin{table}[h!]
\resizebox{\linewidth}{!}{
    \centering
        \begin{tabular}{c|c|cc|cc}
        \hline
        \multirow{2}{*}{\textbf{Dataset}} & \multirow{2}{*}{\textbf{Size}} & \multicolumn{2}{c|}{\textbf{Llama-3.1}} &  \multicolumn{2}{c}{\textbf{Qwen2} }\\
        \cline{3-6}
        & &\textbf{Baseline} & \textbf{Ours} & \textbf{Baseline} & \textbf{Ours} \\
        \hline
        WMT22 & 2037 & 28.10 & 29.32 & 23.09 & 24.98  \\
        WMT23& 557 & 31.01 & 31.83 & 25.99 & 27.37  \\
        WMT24& 998 & 26.09 & 27.46 & 21.82 & 22.43  \\
        \hline
        \end{tabular}
    }
    \caption{The size of WMT testset $EN \to DE$ and the BLEU score using Meta-Llama3.1-Instruct  and Qwen2-7B-Instr}
    \label{tab3}
\end{table}

By incorporating WMT22-24, we ensured that the evaluation process was more robust and free from he potential issues associated with overfitting to previously seen data. 
As shown in Table \ref{tab3}, the results from our experiments using these newer datasets demonstrated a consistent improvement in translation quality across various language pairs. 
The upward trend in performance across all experimental conditions underscores the effectiveness of these datasets in providing a cleaner and more reliable evaluation framework.
This improvement serves to validate validates the robustness of our methodology and also highlights the importance of using up-to-date and diverse data for evaluating LLMs in machine translation tasks.

These findings underscore the importance of addressing data contamination in training large-scale models, as reliance on outdated or overly familiar datasets can obscure the true performance capabilities of the models. 
By employing the latest evaluation datasets, we were able to obtain a more accurate and meaningful assessment of translation quality, thereby providing a clearer picture of the improvements brought about by our approach.

\subsection{Ablation Studies}
\noindent

In order to evaluate the impact of different keyword identification methods, four approaches were compared: (1) translations without word constraints, (2) fixing the number of words extracted from the source sentence, (3) randomly selecting words from the source sentence and (4) using the LLM to identify important words in the source sentence. As shown in Figure \ref{fig3}, our LLM-based approach significantly outperforms both the fixed and random extraction methods, as well as the unconstrained translation. 
This evidence substantiates the assertion that dynamically identifying key terms with LLM guidance and incorporating word constraints is more efficacious strategy for improving translation quality than depending on fixed, random, or unconstrained approaches.

\begin{figure}[h!]
    \centering
    \includegraphics[width=1.0\linewidth]{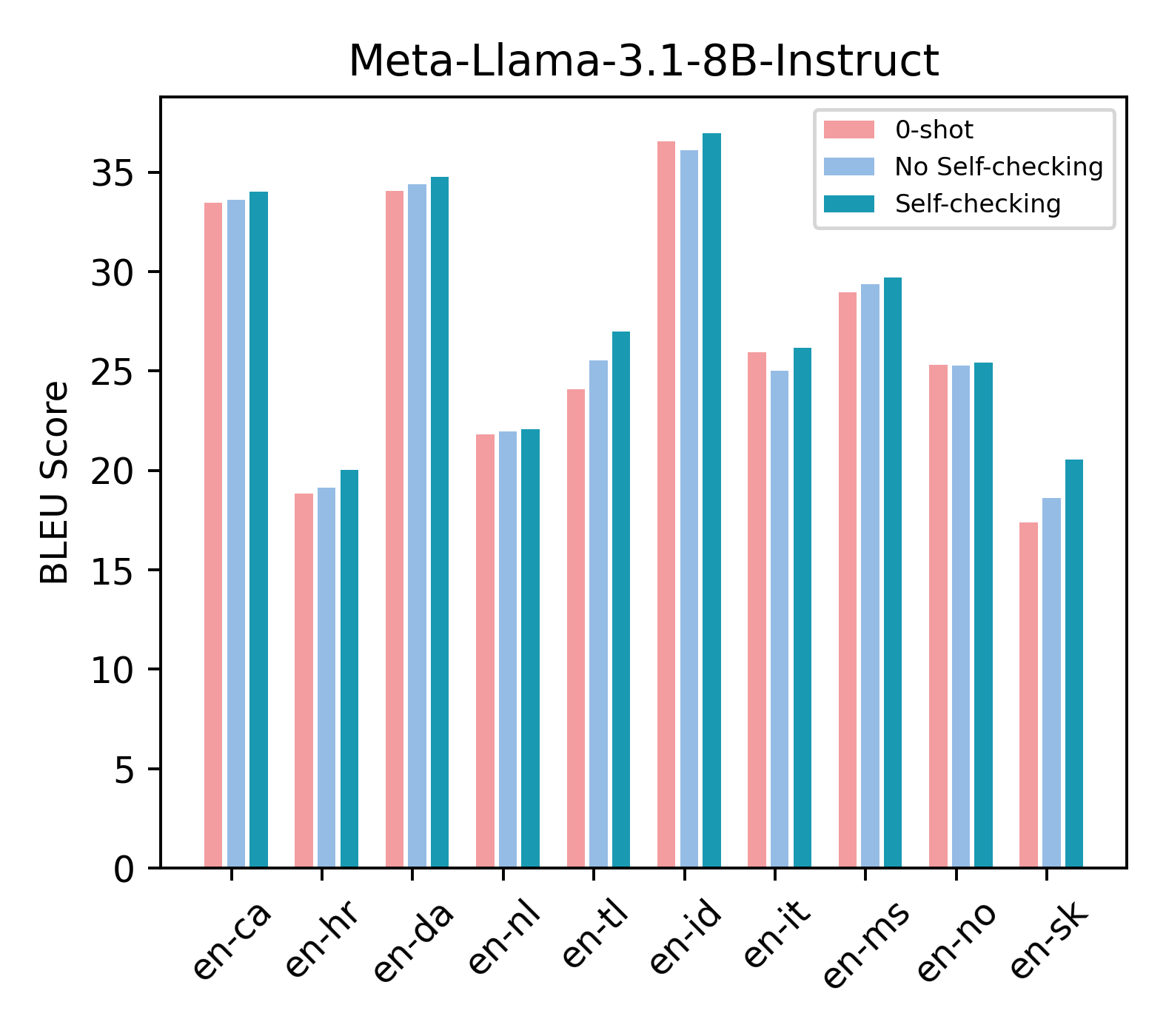}
    \caption{Comparison of BLEU scores for Meta-Llama-3.1-8B-Instruct with and without self-checking versus 0-shot.}
    \label{fig4}
\end{figure}

\begin{figure}[h!]
    \centering
    \includegraphics[width=1.0\linewidth]{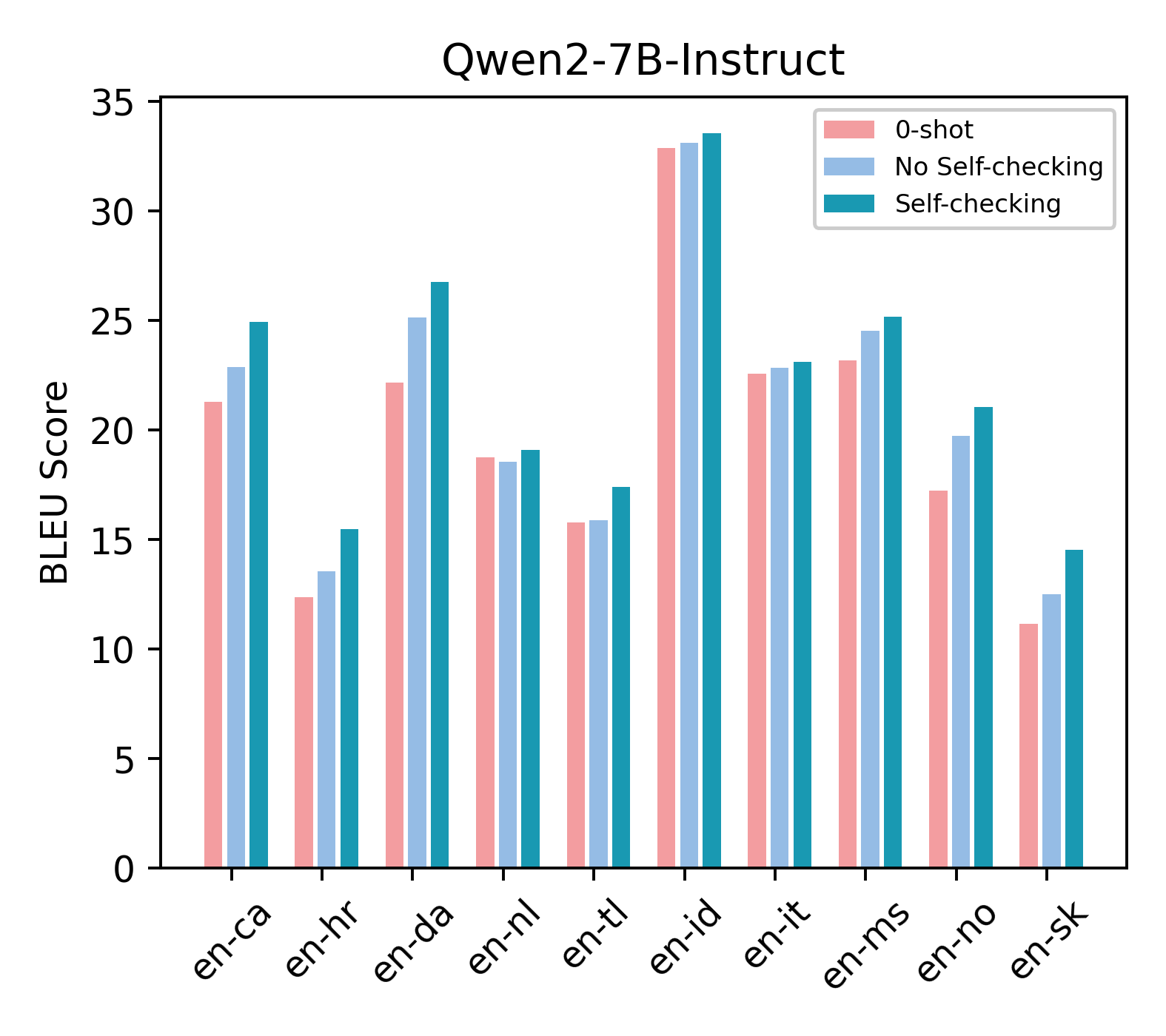}
    \caption{Comparison of BLEU scores for Qwen2-7B-Instruct with and without self-checking versus 0-shot.}
    \label{fig5}
\end{figure}

In scenarios where solely Translation Notes were utilized without the incorporation of Self-checking, we observed a notable enhancement in translation quality when compared to standard zero-shot translations. 
However, the outcomes remained less efficacious than those attained when self-checking was incorporated into the translation procedure.
The absence of Self-checking meant that the model lacked an additional layer of refinement, which is of paramount importance in ensuring that the generated translations adhere to the intended meaning and context.
The results are illustrated in Figures \ref{fig4} and \ref{fig5}.

Overall, these findings provide compelling evidence that our approach is indeed effective.
The utilisation of Translation Notes, particularly when combined with Self-checking, results in enhanced translation quality, while underscoring the necessity for a balanced approach with regard to note length and model capacity.

\section{Conclusion}
\noindent


In this paper, we have presented a novel multi-step prompting approach for enhancing the faithfulness and robustness of LLM-based MT. 
Our method addresses the challenges faced by LLMs in translating rare or specialized terminology by explicitly focusing on key terms in the source sentence and strategically integrating lexical knowledge from high-quality bilingual dictionaries. 
We further leverage the reflective capabilities of LLMs by employing an iterative self-checking mechanism that allows the model to refine its translations based on both lexical and semantic constraints.
Comprehensive experiments conducted on the FLORES-200 benchmark for low-resource languages and contamination-free WMT datasets demonstrate the effectiveness of our approach.

\section{Limitations}
\noindent


One notable limitation of our approach is its reliance on the quality of the bilingual dictionaries used. The effectiveness of the Retrieval-Augmented Translation and Self-checking methods depends on the accuracy and completeness of these dictionaries. Errors, outdated entries, or incomplete translations in the dictionaries can negatively impact the model’s ability to retrieve accurate word translations, thereby affecting overall translation quality. Additionally, dictionary quality may vary across language pairs and domains, leading to inconsistent performance. To address this, dictionaries should be continuously updated and refined. Future work could focus on dynamically improving dictionary quality or integrating multiple translation data sources to alleviate these issues.

\section{Ethical Considerations}
\noindent

Our approach encounters several ethical challenges, primarily concerning the quality and inclusivity of bilingual dictionaries. These dictionaries can introduce biases or inaccuracies, which may be reflected in the translations and potentially perpetuate existing prejudices. Additionally, the reliance on these dictionaries might disadvantage languages with limited representation, leading to unequal access to information. Ensuring data privacy and addressing these biases are crucial to maintaining the fairness and integrity of the translation process. Addressing these concerns proactively is essential to ensure that our method is both responsible and equitable.

\bibliography{custom}




\end{document}